\documentclass[preprint,review,12pt]{elsarticle}
\usepackage{graphicx,multirow}
\usepackage{amsmath,epsfig,amssymb, balance, subfigure, makecell, array, url, algorithmic, algorithm,rotating}
\begin{document}
\begin{frontmatter}
\title{Classifier ensemble creation via false labelling}

\author{B\'alint~Antal}
\address{
Faculty of Informatics, University of Debrecen, 4010 Debrecen, POB. 12, Hungary\protect\\
E-mail: \texttt{antal.balint@inf.unideb.hu}.
}
\begin{abstract}
In this paper, a novel approach to classifier ensemble creation is presented. While other ensemble creation techniques are based on careful selection of existing classifiers or preprocessing of the data, the presented approach automatically creates an optimal labelling for a number of classifiers, which are then assigned to the original data instances and fed to classifiers. The approach has been evaluated on high-dimensional biomedical datasets. The results show that the approach outperformed individual approaches in all cases.
\end{abstract}

\begin{keyword}
Ensemble learning, Diversity, Hidden Markov Random Fields, Simulated annealing, Bioinformatics
\end{keyword}


\end{frontmatter}
\section{Introduction}

Classification is a fundamental task in machine learning. In numerous application fields very complex data needs to be classified which is often a difficult task for a single machine learning classifier \cite{antal_tbme2012} \cite{antal_kbs2014}. There are tremendous amount of research on improving the classification performance in such cases. One highly investigated field for this problem is ensemble learning \cite{kuncheva}, where multiple prediction are fused the produce a more efficient classification approach. One fundamental requirement for the creation of classifier ensembles is diversity among them \cite{Brown2005}, that is, the classifiers included in the ensemble need to complement each other to provide more generalization capabilities than a single learner.  Bagging \cite{bagging} uses randomly selected training subsets with possible overlap (bootstrapping \cite{bootstrap}) to ensure diversity among the member of the ensemble. Other diversity creation techniques may involve disjoint random sampling (random subspace methods \cite{randsub}, for example, some variants of Random Forest algorithms \cite{randomforest}), while Adaboost \cite{adaboost} based techniques aims to increase the accuracy of a weak learner iteratively (boosting \cite{boosting}) using targeted sampling: each iteration considers the misclassified instances of the training data to be more important, and drives the iteration process to include them in the current training set.  Another approach to create diverse ensembles is ensemble selection \cite{Ruta}, where diversity of classifiers trained on the same dataset is measured and an optimal subset is selected.    

A more comprehensive review on the above described techniques can be found in \cite{polikar}. 
The relationship of classifier diversity and ensemble accuracy is highly investigated in the ensemble learning community. Although the definite connection between diversity measures and ensemble accuracy is an open question \cite{measures}, a decomposition of majority voting error into good and bad diversity is proposed in \cite{BrownMCS10}. 

\begin{figure*}
\centering
\includegraphics[width=\linewidth]{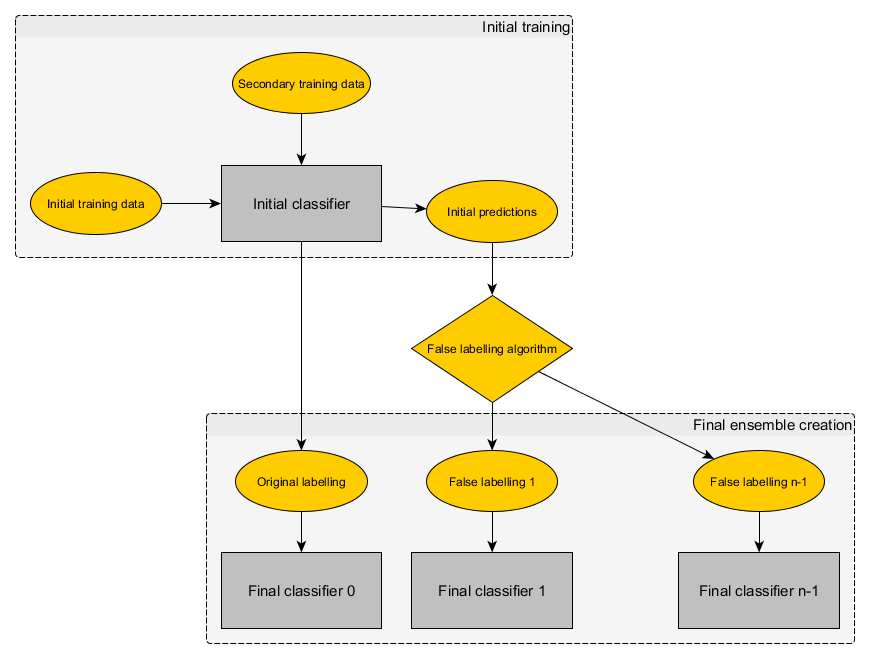}
\caption{Flowchart of ensemble creation via false labelling}
\label{fig:false} 
\end{figure*}

In this paper, a novel approach for ensemble creation based on this theoretical result is presented. The proposed approach takes the predictions of a single classifier on a training set. Then, an optimal labelling complimenting the predictions of the classifiers are created. Thus, an optimal but false labelling set is created for a number of classifiers. The data with each false labelling is trained to a classifier thus forming an ensemble. We define a Markov Random Field problem to create an optimal ensemble with this method. The approach has been tested on high-dimensional biomedical datasets where a large improvement over a single learner is achieved. Other aspects of the algorithm including its performance comparison with different number of ensemble members is also discussed. The outline of the proposed algorithm can be seen in Figure \ref{fig:false}.

The rest of the paper is organized as follows: section \ref{sec:proposed_method} contains the mathematical background behind the proposed method, while section \ref{sec:optimization} defines an optimization problem to solve it and proposes an implementation for it. Section \ref{sec:methodology} contains our experimental details, while the results are presented and discussed in section \ref{sec:results}. Finally, conclusions are drawn in section \ref{sec:conclusion}.

\section{Ensemble creation via false labelling}
\label{sec:proposed_method}

The presented false labelling based ensemble creation are presented is restricted to binary classification problems.  In this section, the mathematical background behind the algorithm is presented. Moreover, an optimization problem is defined to provide an efficient solution for the false labelling problem.
For the basic machine learning and ensemble definitions, we relied on the classic literature \cite{kuncheva} and \cite{BrownMCS10}.

\newcounter{qc}
Let $\Omega = \left\{ -1, +1\right\}$ be a set of class labels. Then, a function 
\begin{equation}
D: \mathbb{R}^{n} \rightarrow \Omega
\label{eq:classifier}
\end{equation} 
is called a classifier, while a vector $\vec{\chi} = \left(\chi_{1}, \chi_{2}, \dots, \chi_{n}\right) \in \mathbb{R}^{n}$ is called a feature vector. A dataset $T \in \{R^{n} \times \Omega\}^{l}$ can be defined as follows: 
\begin{equation}
T = \{\langle\vec{\chi}_{0}, \omega_{0}\rangle, \langle\vec{\chi}_{1}, \omega_{1}\rangle, \dots, \langle\vec{\chi}_{k}, \omega_{k}\rangle\},
\label{eq:dataset}
\end{equation}
where  $\vec{\chi}_{i} \in \mathbb{R}^{n}, \omega_{k} \in \Omega, i = 1,\dots,k$ are feature vectors and labels, respectively. 

Let $D_{1}, D_{2}, \dots, D_{L}$ be classifiers and $d_{t}\left(\vec{\chi}\right) \in \Omega,\,t=1,\dots,L$ their output on the feature vector $\vec{\chi}$. Then, the output of the majority voting ensemble classifier ${\cal D}_{maj} : \mathbb{R}^{n} \rightarrow \Omega$ can be defined as follows:
\begin{equation}
d_{maj} \left(\vec{\chi}\right) = sign\left(\dfrac{1}{L}\sum_{t=1}^{L} d_{t}\left(\vec{\chi}\right)\right).
\label{eq:majority}
\end{equation}


The creation of an ensemble ${\cal D}_{maj}$ of $L$ classifiers (equation \ref{eq:classifier}) starts by training a base classifier on the half of the training dataset (equation \ref{eq:dataset}) $T$ ($T_{0}$). We take the output $C_{orig}$ of the classifier $D_{orig}$ on the other half of the training set ($T_{1}$) and create $L-1$ optimal labellings for a the remaining base classifiers $D_{i},i = 2,\dots,L$. Then, we train these classifiers on $T_{1}$ with their respective false labellings ${\cal C}_{false}^{i}$.

The outline of the ensemble creation method is summarized in algorithm \ref{alg:false}, while the mathematical formulation is presented in the rest of the section.

\begin{algorithm}
\caption{Outline of ensemble creation via false labelling}
\label{alg:false}
\begin{algorithmic}[1]
	\REQUIRE a dataset $T \neq \emptyset$, a label set ${\cal C} \neq \emptyset$, a classifier $D_{orig}$, the number of ensemble members $L > 2$ ($L$ is odd).
	\ENSURE an ensemble of trained classifiers ${\cal D}_{maj}$.
	\STATE Split $T$ into $T_{0}$ and $T_{1}$ randomly.
	\STATE Train $D_{orig}$ on $T_{0}$.
	\STATE $C_{orig} \leftarrow D_{orig}\left(T_{1}\right)$
	\STATE $C_{cl} \leftarrow F\left(C_{orig}\right) = \{{\cal C}_{false}^{2}, {\cal C}_{false}^{3}, \dots, {\cal C}_{false}^{L}\}$	
	\FOR{$i \leftarrow 2, \dots, L$}  
		\STATE Train a classifier $D_{i}$ on $LC\left(T_{1}, {\cal C}_{false}^{i}\right), {\cal C}_{false}^{i} \in C_{cl}$.
	\ENDFOR
	\RETURN $\{D_{orig}, D_{2}, \dots, D_{L}\}$ 
 \end{algorithmic}
\end{algorithm}

\subsection{Ensemble creation}

The proposed ensemble creation depends on the output of one classifier $D_{orig}$ for a given training dataset $T$.

First, we split $T$ into two equal parts $T^{\left(0\right)}$ and $T^{\left(1\right)}$ randomly. We train $D_{orig}$ on $T^{\left(1\right)}$ and classify all $\vec{\chi}^{1}_{j} \in T^{\left(0\right)}, j =1,\dots, k/2$ element of $T^{\left(1\right)}$:
\begin{equation}
	{\cal C}_{orig}^{1} = \{\omega_{j}|
	\omega_{j} = D_{orig}\left(\vec{\chi}^{1}_{j}\right), \vec{\chi}^{1}_{j} \in T^{1}, j =1,\dots, k/2
	\}.
\end{equation}
Then, we create a majority voting classifier ensemble of $L$ members:
\begin{equation}
	{\cal D}_{maj} = \{D_{1} = D_{orig}, D_{2}, \dots, D_{L}\}.
\end{equation}
To train $D_{2}, \dots, D_{L}$, we will define a false labelling function $F: \Omega^{k/2} \rightarrow \Omega^{k/2 \cdot \left(L-1\right)}$. That is
\begin{equation}
	F\left({\cal C}_{orig}^{1}\right) = \{{\cal C}_{false}^{2}, {\cal C}_{false}^{3}, \dots, {\cal C}_{false}^{L}\},
\label{eq:fl}
\end{equation}
where ${\cal C}_{false}^{i} = \{\omega^{f}_{i,j}|
	\omega^{f}_{i,j} \in \Omega, i = 2,\dots, L, j =1,\dots, k/2.
	\}$.
To apply the new labels to the existing dataset, we define the label changing operation $LC: \{R^{n} \times \Omega \times \Omega\}^{l} \rightarrow \{R^{n} \times \Omega\}^{l}$ in the following way:
\begin{equation}
	LC\left(T, {\cal C}\right) = \{\langle \vec{\chi}_{j}, \omega^{f}_{j} \rangle | \langle \vec{\chi}_{j}, \omega_{j} \rangle \in T, \omega^{f}_{j} \in {\cal C} \},
\end{equation}
where $T$ is a dataset and ${\cal C}$ is a label set.
Finally, we train $D_{i}, i = 1, \dots, L$ on $LC\left(T, {\cal C}_{false}^{i}\right)$, where ${\cal C}_{false}^{i} \in F\left({\cal C}_{orig}^{1}\right)$. Then, the false labelling ensemble is created.

\subsection{Selection of the false labelling function}

To define an optimal false labelling function $F$ (see equation \ref{eq:fl}), we recite the decomposition of the majority voting error described in \cite{BrownMCS10}. The majority voting error can be split into three terms: the individual error of the classifiers , the disagreement of the classifiers when they classified the input correctly ("good diversity") and the disagreement of the classifiers when they classified the input incorrectly ( "bad diversity"). The majority voting error decomposition is the basis for defining the energy function for our method.

Let $y\left(\vec{\chi}\right)$ be the true class label for the feature vector $\vec{\chi}$. Then, the zero-one loss for
$d_{t}\left(\vec{\chi}\right)$ is defined as follows \cite{BrownMCS10}:
\begin{equation}
e_{t}\left(\vec{\chi}\right) = \dfrac{1}{2} \left(1 - y\left(\vec{\chi}\right)d_{t}\left(\vec{\chi}\right) \right)
\end{equation}
Then, the average individual zero-one loss is \cite{BrownMCS10}
\begin{equation}
e_{ind}\left(\vec{\chi}\right) = \dfrac{1}{L} \sum_{t=1}^{L} e_{t}\left(\vec{\chi}\right)
\end{equation}
and the ensemble zero-one loss is:
\begin{equation}
e_{maj}\left(\vec{\chi}\right) = \dfrac{1}{2} \left(1 - y\left(\vec{\chi}\right)d_{maj}\left(\vec{\chi}\right) \right)
\label{eq:majerr}
\end{equation}
 
The disagreement between $d_{t}$ and the ensemble is the following \cite{BrownMCS10}:
\begin{equation}
\delta_{t}\left(\vec{\chi}\right) = \dfrac{1}{2} \left(1 - d_{t}\left(\vec{\chi}\right)d_{maj}\left(\vec{\chi}\right) \right).
\label{eq:dis}
\end{equation}
%
The classification error of an ensemble is defined \cite{BrownMCS10} as follows: 
\begin{equation}
E_{maj} = \int_{\vec{\chi}} e_{ind} -  \int_{\vec{\chi^{+}}} \dfrac{1}{L} \sum_{t=1}^{L} \delta_{t}\left(\vec{\chi}\right) + \int_{\vec{\chi^{-}}} \dfrac{1}{L} \sum_{t=1}^{L} \delta_{t}\left(\vec{\chi}\right) 
\label{eq:classerr}
\end{equation}

Based on equations \ref{eq:majerr}-\ref{eq:classerr}, an optimization problem can be defined to find such an optimal labelling. 
\section{Optimization via Hidden Markov Random Fields}
\label{sec:optimization}

To solve the optimization problem, an approach based on Hidden Markov Random Fields (HMRF) in presented. HMRF is a powerful framework for solving large-scale optimization problems, since there are multiple methods for solving HMRF problems near optimally in normal time, which would be a challenging task to find exact false labellings for real-life applications.

In this section, we briefly summarize the basis for Hidden Markov Random Field (HMRF) optimization based on \cite{markov}.
Let 
\[
	A_{k/2,L-1} = a_{i,j} = 
 \begin{pmatrix}
  \omega^{f}_{1,1} & \omega^{f}_{1,2} & \cdots & \omega^{f}_{1,k/2} \\
  \omega^{f}_{2,1} & \omega^{f}_{2,2} & \cdots & \omega^{f}_{2,k/2} \\
  \vdots  & \vdots  & \ddots & \vdots  \\
  \omega^{f}_{L-1,1} & \omega^{f}_{L-1,2} & \cdots & \omega^{f}_{L-1,k/2} \\
 \end{pmatrix}
\]
be a matrix containing a false labelling setup and ${\cal C}_{orig} = b_{i,j}$
a vector containing the labellings of $D_{orig}$ and ${\cal C}_{training} = c_{i,j}$ the labels assigned originally the training instances. All $a_{i,j}$ is a variable which can contain a possible label and at the end of the optimization process, each row contain a false labelling for a classifier $D_{i}$. 
  
Let $\Lambda = \{0, 1\}$ be a set of labels. Then, we assign each $a_{i,j}, i=1,\dots,k/2j=1,\,\dots,\,L-1$ a label $\omega_{i_{j}}$. Let $X$ be a labelling field. $X$ is a Markov Random Field if $P\left(X = \omega\right)$, for all $\omega \in \Lambda$ and $P\left(\omega_{a_{i,j}} | \omega_{a_{k,l}},\, a_{i,j} \neq i_{k} \right) = P\left(\omega_{a_{i,j}} | \omega_{a_{k,l}},\, a_{k,l} \in N_{a_{i,j}}\right)$, where $N_{a_{i,j}}$ is a neighbourhood of $a_{i,j}$.

The optimal labelling for the $A$ variables with the HMRF optimization, one can use the the Hammersley-Clifford Theorem \cite{hct} to calculate the global energy for a labelling by summarizing the local energies for each variable. That is, during the optimization process, the global energy would be a function of the changes in the states of the $a_{i,j}$ variables. 

We define the following three neighbourhoods for the optimization process: 
\begin{equation}
	N_{a_{i,j}}^{1} = \{a_{m,j} | m \in \{1, k/2\}, m \neq i\} \cup \{b_{i}\},
\end{equation} 
and a neighbourhood of a single variable containing the labelling for all of the feature vectors for the same classifier 
\begin{equation}
	N_{a_{i,j}}^{2} = \{a_{i,l} | l \in \{1, L-1\}, l \neq j\} \cup \{b_{i}\},
\end{equation} 
which is a neighbourhood of a single variable containing the labelling of the other classifiers for the same feature vector,
and
\begin{equation}
	N_{a_{i,j}}^{3} = \{a_{k,l} | k \in \{i-q, i+q\}, l \in \{j-q, j+q\}\},
\end{equation}
which is a neighbourhood of a variable containing labelling of its close classifiers for inputs in a $q \cdot q$ part of $A$.
First, we consider the individual classification error the individual classifiers:
\begin{equation}
	U_{ind}\left(a_{i,j}\right) = \dfrac{\sum\{a_{k,l} | a_{k,l} \in N_{a_{i,j}}^{1} \wedge a_{k,l} = \omega_{i}\}}{k/2},
\end{equation}
where $\omega_{i}$ is the actual label assigned to the feature vector in the training set.
Out next criteria for the optimization process is to give a labelling, where the number of correct votes is exactly 50\%+1 in all cases. Let 
\[
	o = L/2 + 1.
\]
Then, we define the function $E_{votes}$ in the following way:
\begin{equation}
	U_{good}\left(a_{i,j}\right) = \dfrac{\sum\{a_{k,l} | a_{k,l} \in N_{a_{i,j}}^{2} \wedge a_{k,l} = b_{i}\}  - o}{ o}.
\end{equation}
That is, we sum the correct labellings for a given input and subtracting the optimal number of votes from it. In this way, the $E_{votes}$ will be minimal if the number of correct votes is less than or equal to the number of optimal votes. Thus, we maximize the disagreement for bad diversity and minimize to good diversity \cite{BrownMCS10}. To ensure classification accuracy (and avoid having lower numbers of votes resulting from negative values of $E_{votes}$), we also define   
\begin{equation} 
	U_{bad}\left(a_{i,j}\right) = - 
	\dfrac{
		\sum
			\{a_{k,l} | a_{k,l} \in N_{a_{i,j}}^{3}\} 
			\wedge
			\{ a_{k,l} \neq b_{i}\}
			 - o}{o},
\end{equation}
which is the disagreement term for bad diversity.

Finally, we must ensure that the votes are unevenly distributed among the classifiers to have less correlation between variables:
\begin{equation}
	U_{smoothness}\left(a_{i,j}\right) = \begin{cases}
						\beta & if a_{i,j} = a_{k,l}\\
						-\beta & otherwise.
					\end{cases},
\end{equation}
for all $a_{k,l} \in N_{a_{i,j}}^{2}$. In this way we ensure low correlation between the label sets assigned to the classifiers.
In summary, the global energy $U$ is the following:
\begin{equation}
	\begin{split}
	U = 
	\sum_{i = 0}^{k/2}\sum_{j = 0}^{L-1} E_{ind}\left(a_{i,j}\right) + E_{good}\left(a_{i,j}\right) + E_{bad}\left(a_{i,j}\right) +\\ E_{smoothness}\left(a_{i,j}\right).
\end{split}
\end{equation} 
The optimization of the HMRF configuration can be done by optimizing $U$. Since  simulated annealing \cite{sa}, an efficient algorithm for finding approximate global solutions for large state-spaces.

In summary, simulated annealing measures energy values from different states of the variables. Each state is accepted as a better solution if provided a more optimal energy value or accepted by a function, which uses a random number to decide it. This step is important in avoiding stuck in local optima, as do other stochastic approaches like stochastic gradient search. The algorithm for simulated annealing can be found in algorithm \ref{alg:sa}.

\begin{algorithm}[H]
\caption{Solving the optimization problem with simulated annealing.}	
\label{alg:sa}
\begin{algorithmic}[1]
	\REQUIRE An initial temperature $T$, a minimal temperature $T_{min}$ and a temperature change quotient $q$.
	\REQUIRE A function $changeState$ changing variable values from their current state.
	\REQUIRE An acceptance function $accept$. E.g. 
	\begin{equation}		
		accept\left(u,\, u_{best},\, T,\right) = 
 	\begin{cases}
 		true, & \exp{\left(\dfrac{e - e_{i}}{T}\right)} > r,\\
 		false, & \mbox{otherwise},\\
 	\end{cases}
\end{equation} 	
 	where $r$ is a random number.
	\ENSURE An approximation of the optimal false labelling.
	\STATE $A = a_{m,n} \leftarrow \{0\}.$
	\STATE $u \leftarrow U(A)$
	\STATE $l_{best} \leftarrow A$
	\STATE $u_{best} \leftarrow u$
	\STATE $s \leftarrow 0$
	\WHILE{$T \geq T_{min}$}
		\STATE $A \leftarrow changeState\left(A\right)$
		\STATE $u \leftarrow U(A)$
		\IF{$u \geq u_{best}$ or $accept(u, u_{best}, T)$}
			\STATE $l_{best} \leftarrow A$
			\STATE $u_{best} \leftarrow u$		
		\ENDIF
		\STATE $T \leftarrow T \cdot q$	
	\ENDWHILE 
	\RETURN $l_{best}$
\end{algorithmic}
\end{algorithm}

After the optimization process, the $l_{best}$ state of $A$ is the optimal false labelling, which can be used to train the classifiers.  

\section{Methodology}
\label{sec:methodology}

The proposed approach has been evaluated on high-dimensional biomedical datasets containing gene expressions or proteomics data downloaded from the the Keng Ridge repository \cite{keng}. The description of the datasets including the number of instances, the number of features per instance and the status of the patient by disease  is summarized in Table \ref{tab:datasets}. As it can be seen, the datasets contain a large number of features for a small number of instances thus making it challenging classification problems. Thus, the datasets are bootstrapped for training to ensure the number of instances per class are similar for better comparison of the methods.

\begin{sidewaystable}\footnotesize
\renewcommand{\arraystretch}{1.3}
\caption{Description of the datasets}
\label{tab:datasets}
\centering
\begin{tabular}{l c c l}
\hline

Dataset & Number of Instances & Number of Features & Disease\\
\hline
breastCancer-train & 73 & 24481 & \multirow{2}{*}{Breast cancer.}\\
breastCancer-test & 19 & 24481 &\\
\hline 
centralNervousSystem & 60 & 7129 & Central nervous system embryonal tumor. \\
\hline
colonTumor & 62 & 2000 & Colon tumor. \\
\hline
DLBCL-Stanford & 47 & 4026 & \multirow{5}{*}{Diffuse large B-cell lymphoma}\\
DLBCLOutcome & 58 & 6817 & \\ 
DLBCLTumor & 77 & 6817 & \\
DLBCL-NIH-train & 160 & 7399 & \\
DLBCL-NIH-test &  80 & 7399 & \\
\hline
OC0-9 & 216 & 37340 & Ovarian cancer\\
\hline
prostate-tumorVSNormal-train & 102 & 12600 & \multirow{3}{*}{Prostate cancer}\\
prostate-tumorVSNormal-test & 33 & 12600 & \\
prostate-outcome & 21 & 12600 & \\
\hline
\end{tabular}   
\end{sidewaystable}

The datasets were splitted into two equal partitions randomly 10 times to have a fair comparison. The false-labelling ensembles are tested with 3, 5, 7, 9, 11, 13, 15 members with Naive Bayes \cite{naivebayes} base classifiers for each problem. The implementation of the classifiers was done using Weka \cite{weka}. To measure the accuracy of the ensembles, the classification accuracy of each cross-validation round is measured and their mean and standard deviation is calculated. 
For a comparison, we also show the results for a Naive Bayes classifier, which serves base classifiers in the ensembles, and three state-of-the-art ensemble approaches, namely Adaboost, Bagging and Random Forest. 

\section{Results and discussion}
\label{sec:results}

\begin{figure*}
	\centering
	\subfigure[Accuracy] {
		\includegraphics[width=.5\linewidth]{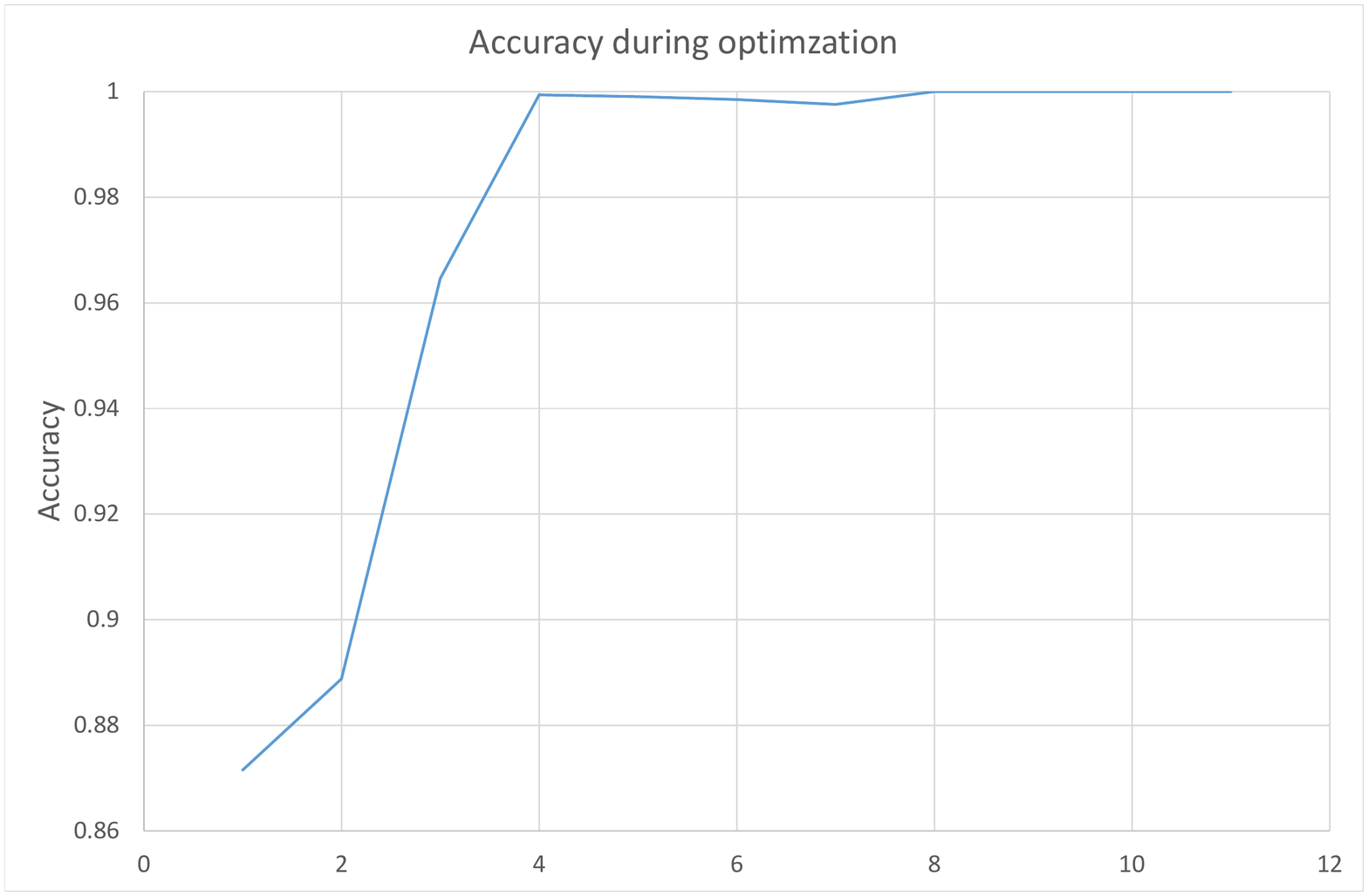}
	\label{fig:opt_acc}	
	}
	\\
	\subfigure[Correlation] {
		\includegraphics[width=0.5\linewidth]{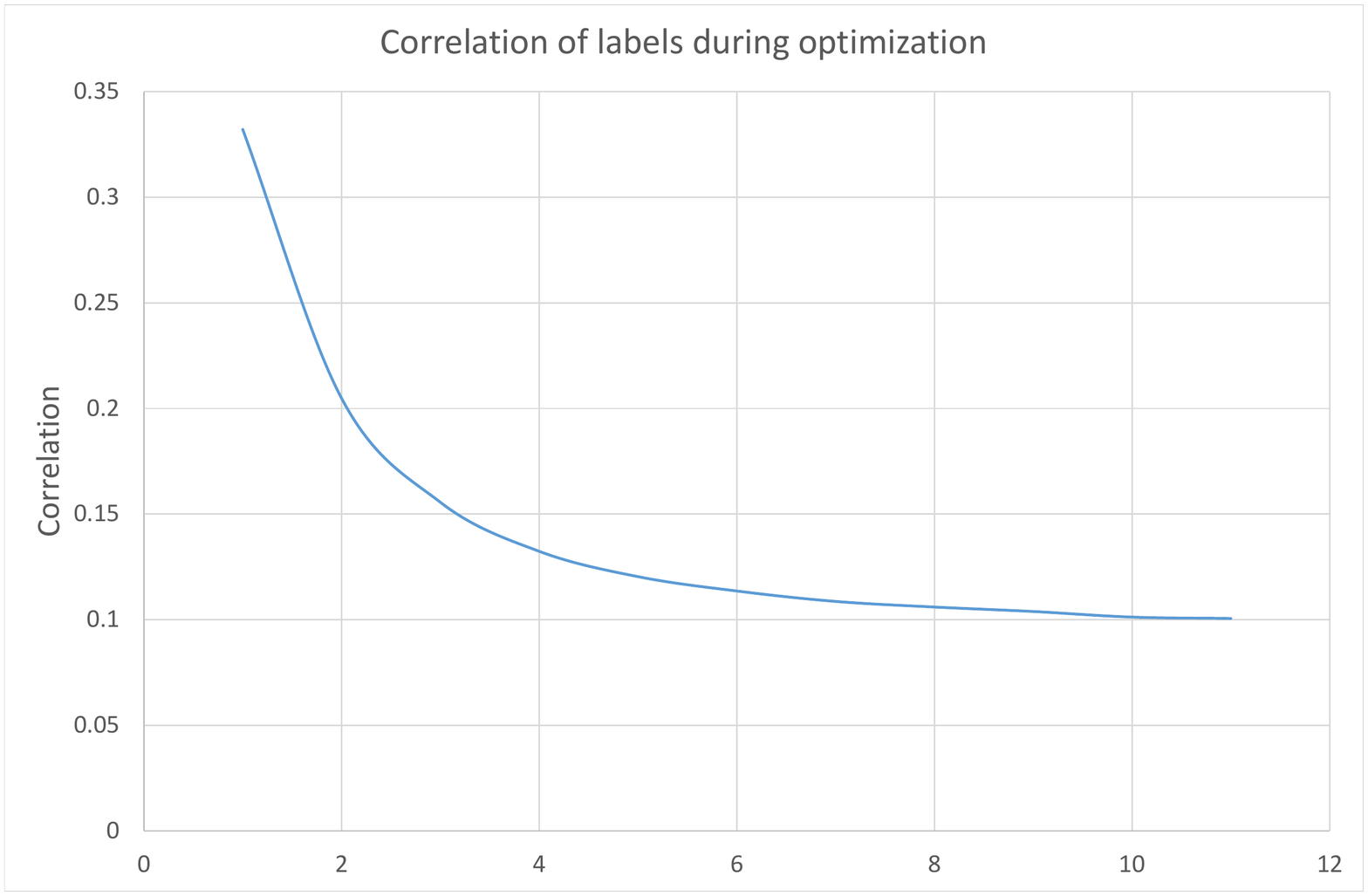}
	\label{fig:opt_corr}	
	}
	\\
	\subfigure[Time] {
		\includegraphics[width=0.5\linewidth]{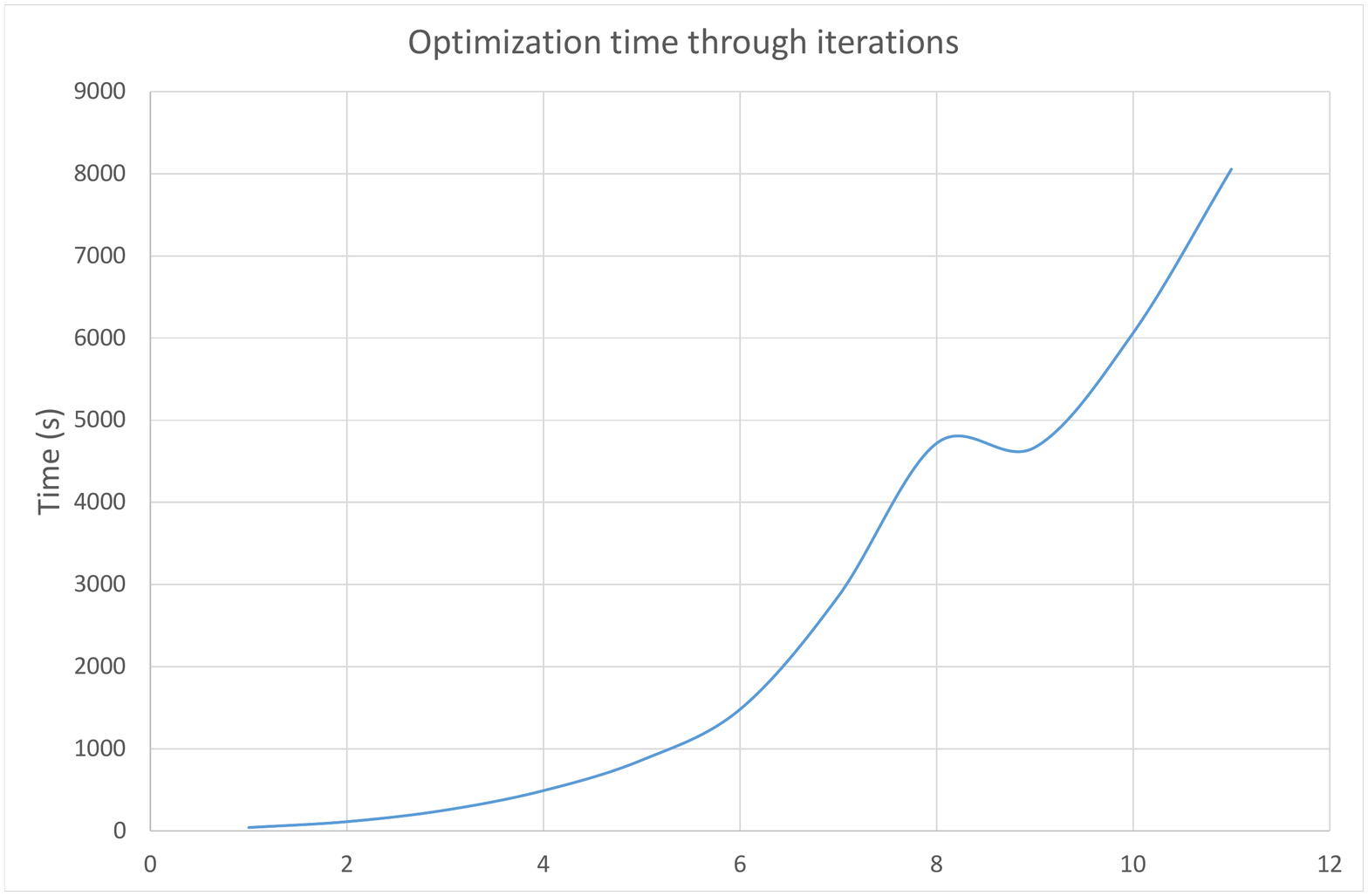}
	\label{fig:opt_time}	
	}
	\label{fig:opt}
	\caption{Accuracy, correlation of the ensemble member labels and execution time through iterations of the optimization procedure}
\end{figure*}

The validity of the optimization technique can be seen in Figures \ref{fig:opt_acc} and \ref{fig:opt_corr}. As it can be seen in this example, the accuracy of the ensemble has increased steadily through iteration converging to an accuracy of 1, while the correlation of the labels of the ensemble members has been decreased at the same time. Figure \ref{fig:opt_time} shows the optimization time through iterations. As it can be seen, in earlier iterations, the optimization procedure increases the energy function with less changes in the labelling spending less time, while in later iterations most of the combinations needs to be tested to increase energy, which require more time.

\begin{figure*}
	 \centering
		\includegraphics[width=\linewidth]{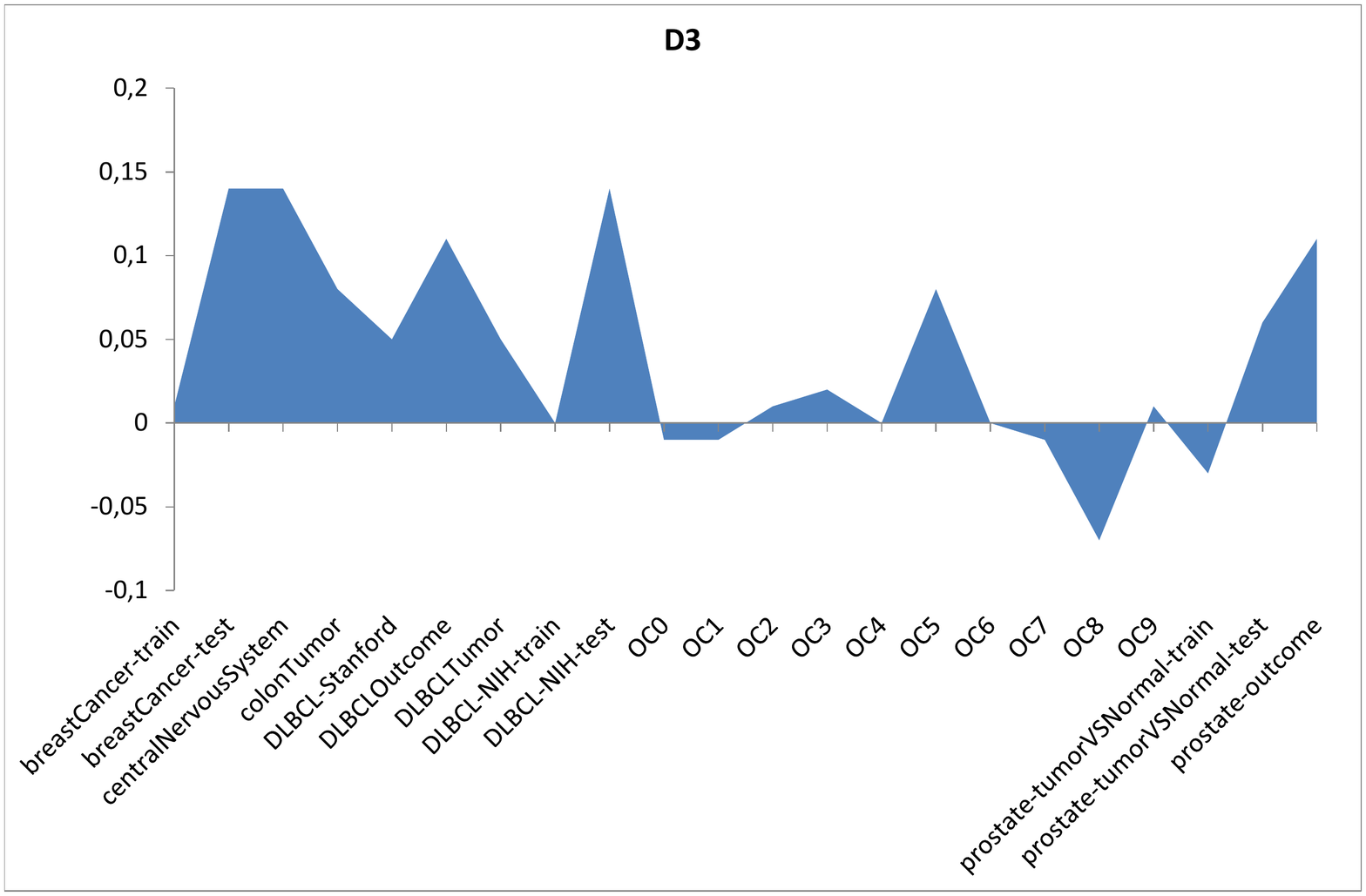}
	\caption{Comparison of the D3 ensemble and the best performing classifiers from Naive Bayes, Adaboost, Bagging and Random Forest.}	
	\label{fig:d3}
\end{figure*}

\begin{figure*}
	 \centering
		\includegraphics[width=\linewidth]{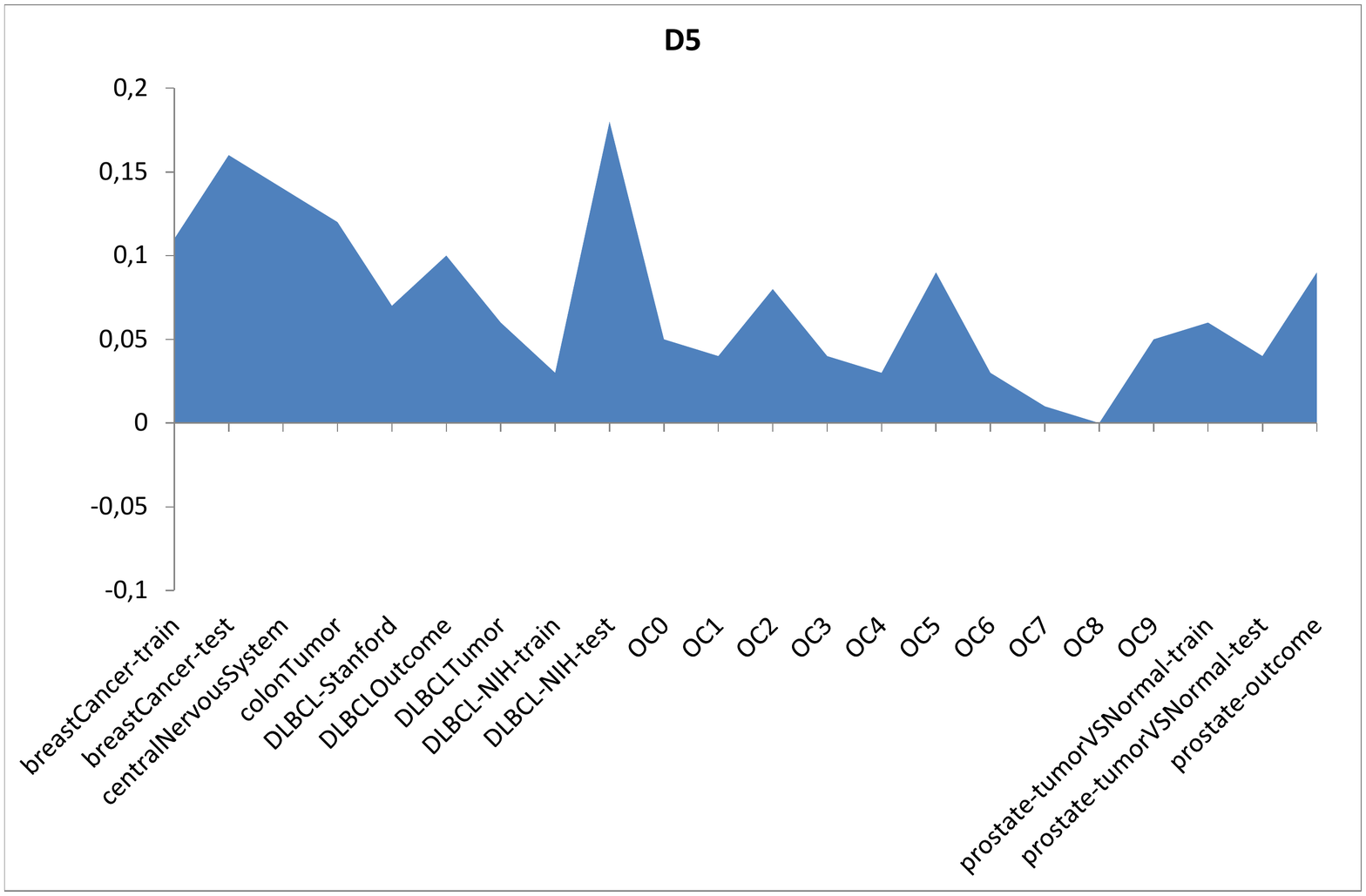}
	\caption{Comparison of the D5 ensemble and the best performing classifiers from Naive Bayes, Adaboost, Bagging and Random Forest.}	
	\label{fig:d5}
\end{figure*}

\begin{figure*}
	 \centering
		\includegraphics[width=\linewidth]{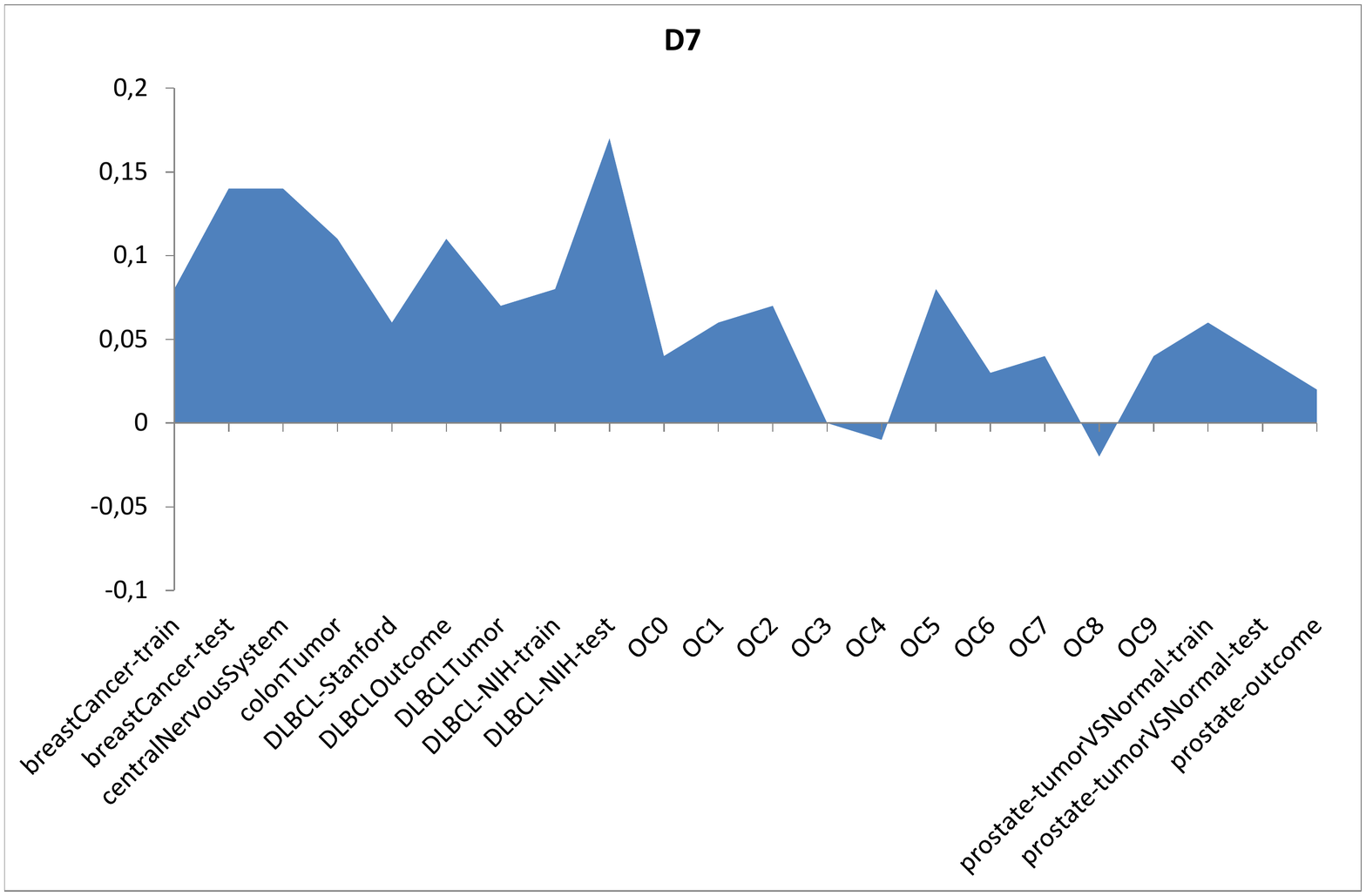}
	\caption{Comparison of the D7 ensemble and the best performing classifiers from Naive Bayes, Adaboost, Bagging and Random Forest.}	
	\label{fig:d7}
\end{figure*}

\begin{figure*}
	 \centering
		\includegraphics[width=\linewidth]{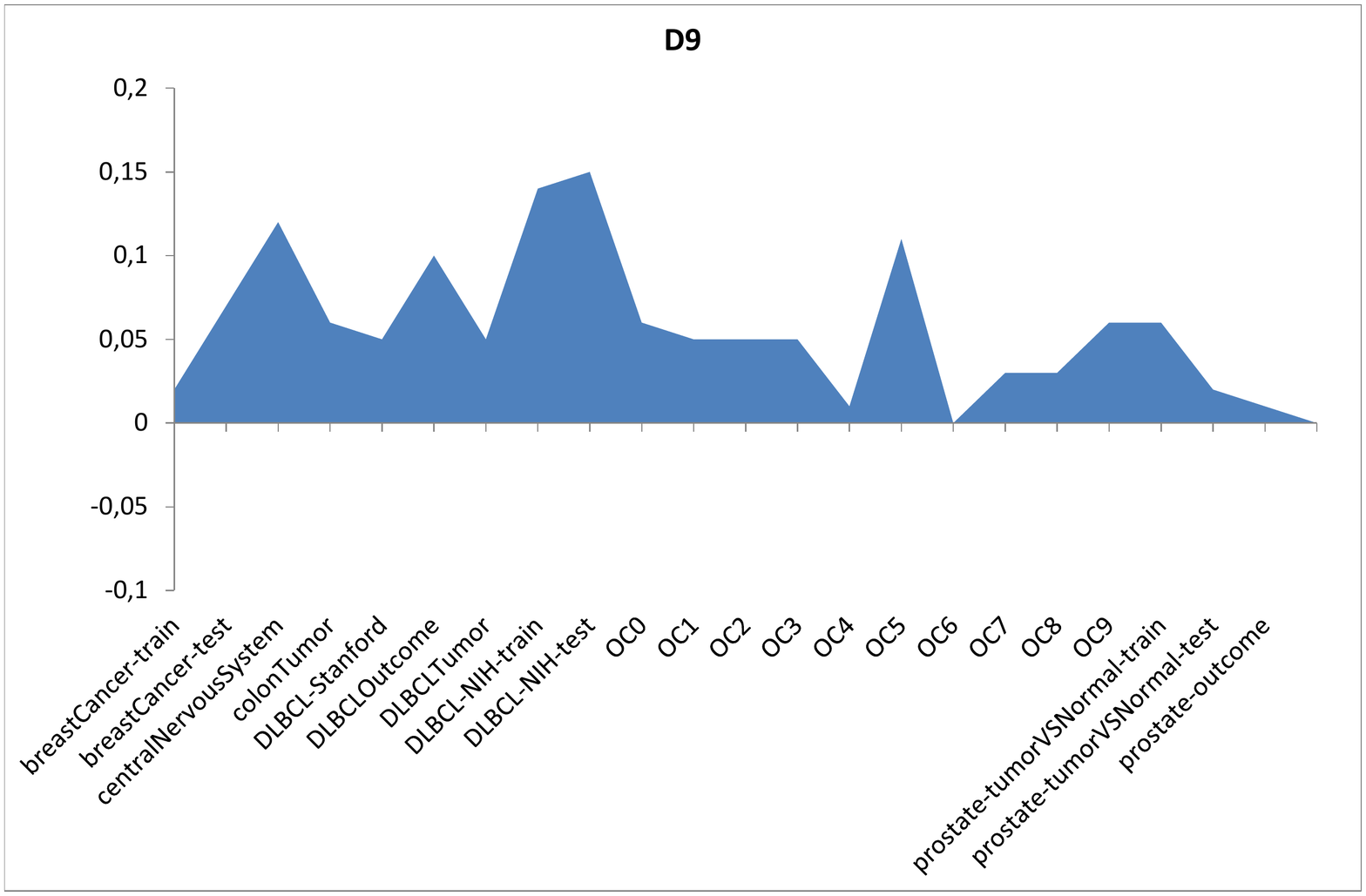}
	\caption{Comparison of the D9 ensemble and the best performing classifiers from Naive Bayes, Adaboost, Bagging and Random Forest.}	
	\label{fig:d9}
\end{figure*}

\begin{figure*}
	 \centering
		\includegraphics[width=\linewidth]{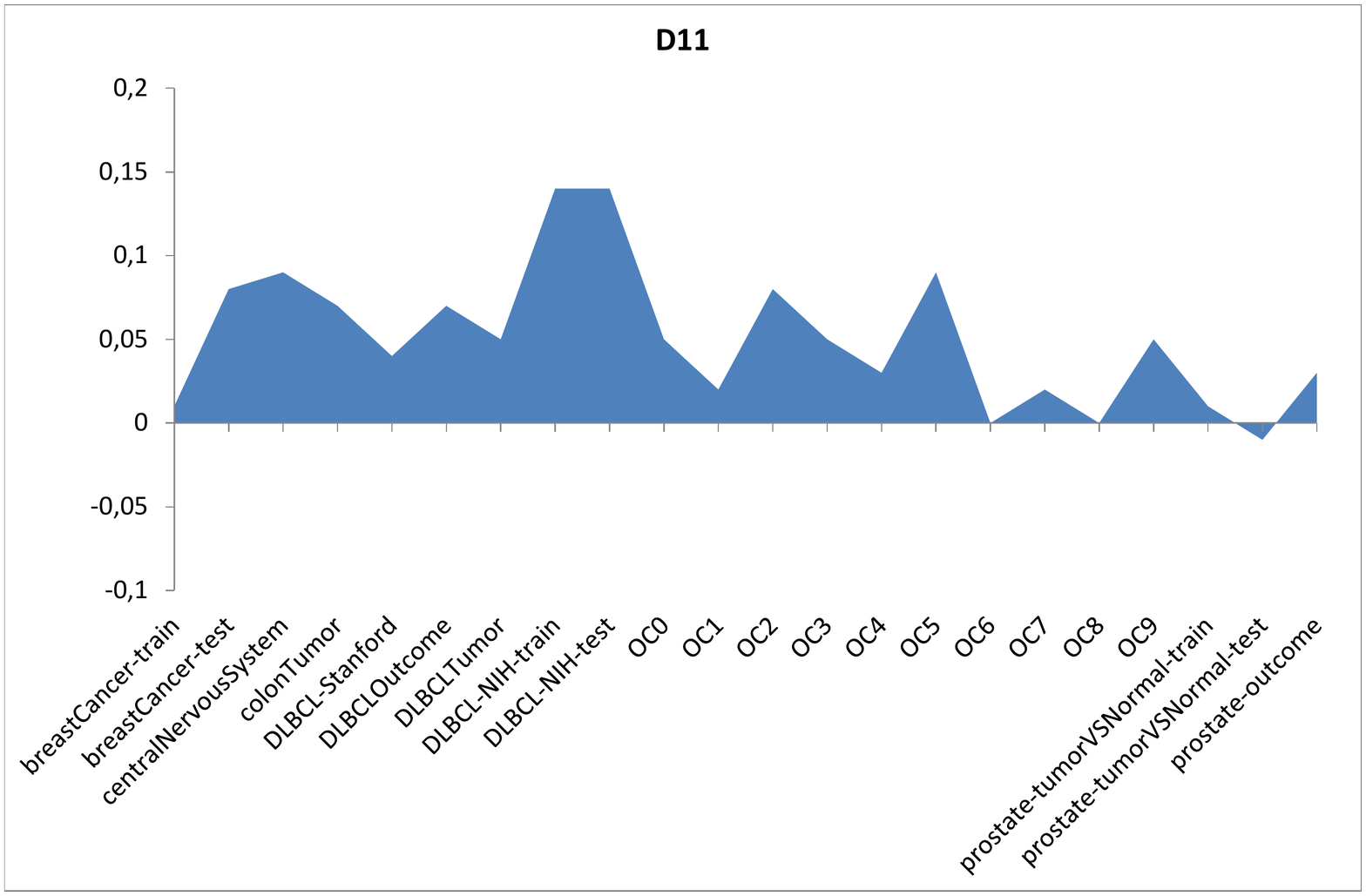}
	\caption{Comparison of the D11 ensemble and the best performing classifiers from Naive Bayes, Adaboost, Bagging and Random Forest.}	
	\label{fig:d11}
\end{figure*}

\begin{figure*}
	 \centering
		\includegraphics[width=\linewidth]{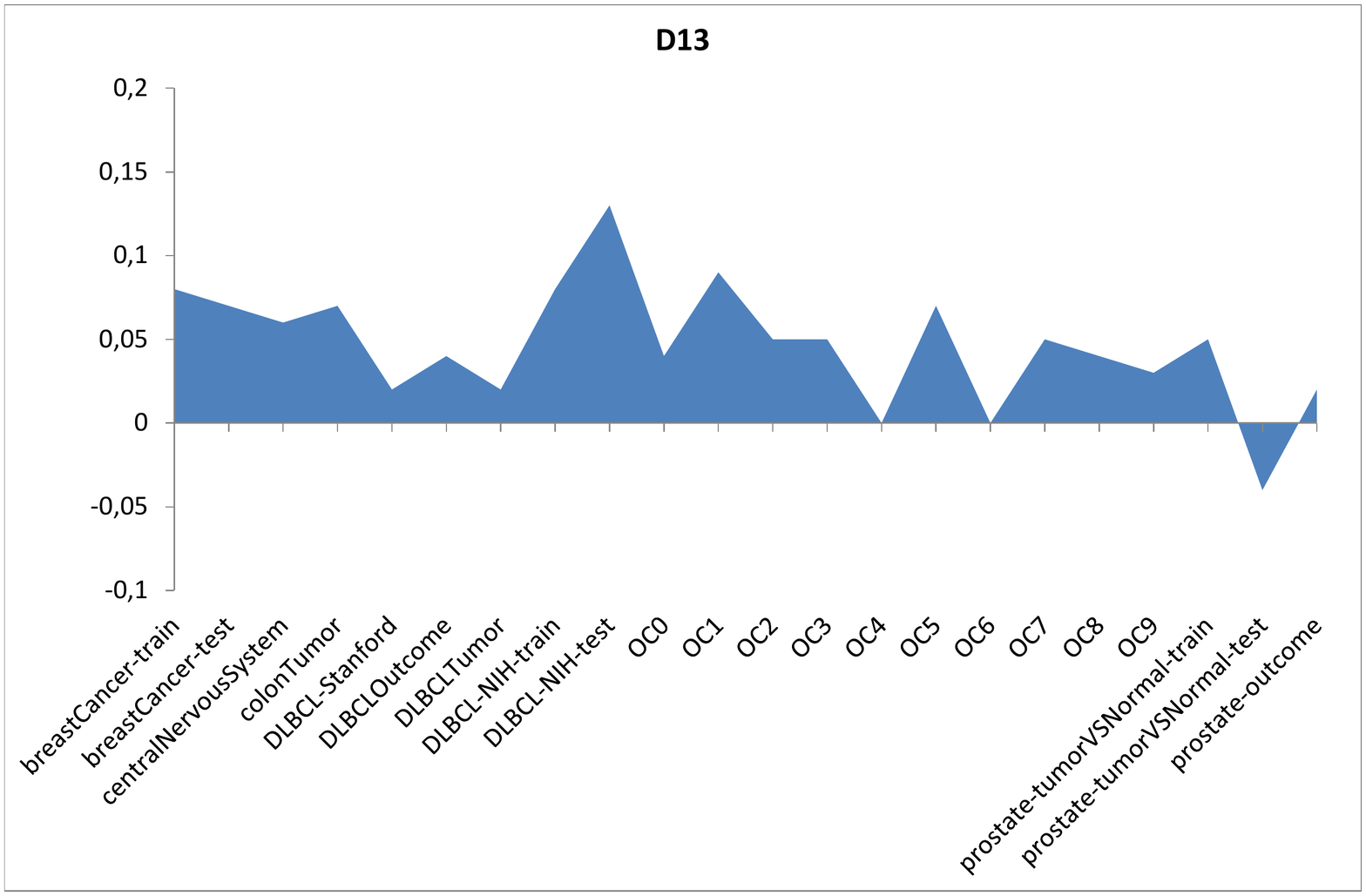}
	\caption{Comparison of the D13 ensemble and the best performing classifiers from Naive Bayes, Adaboost, Bagging and Random Forest.}	
	\label{fig:d13}
\end{figure*}

\begin{figure*}
	 \centering
		\includegraphics[width=\linewidth]{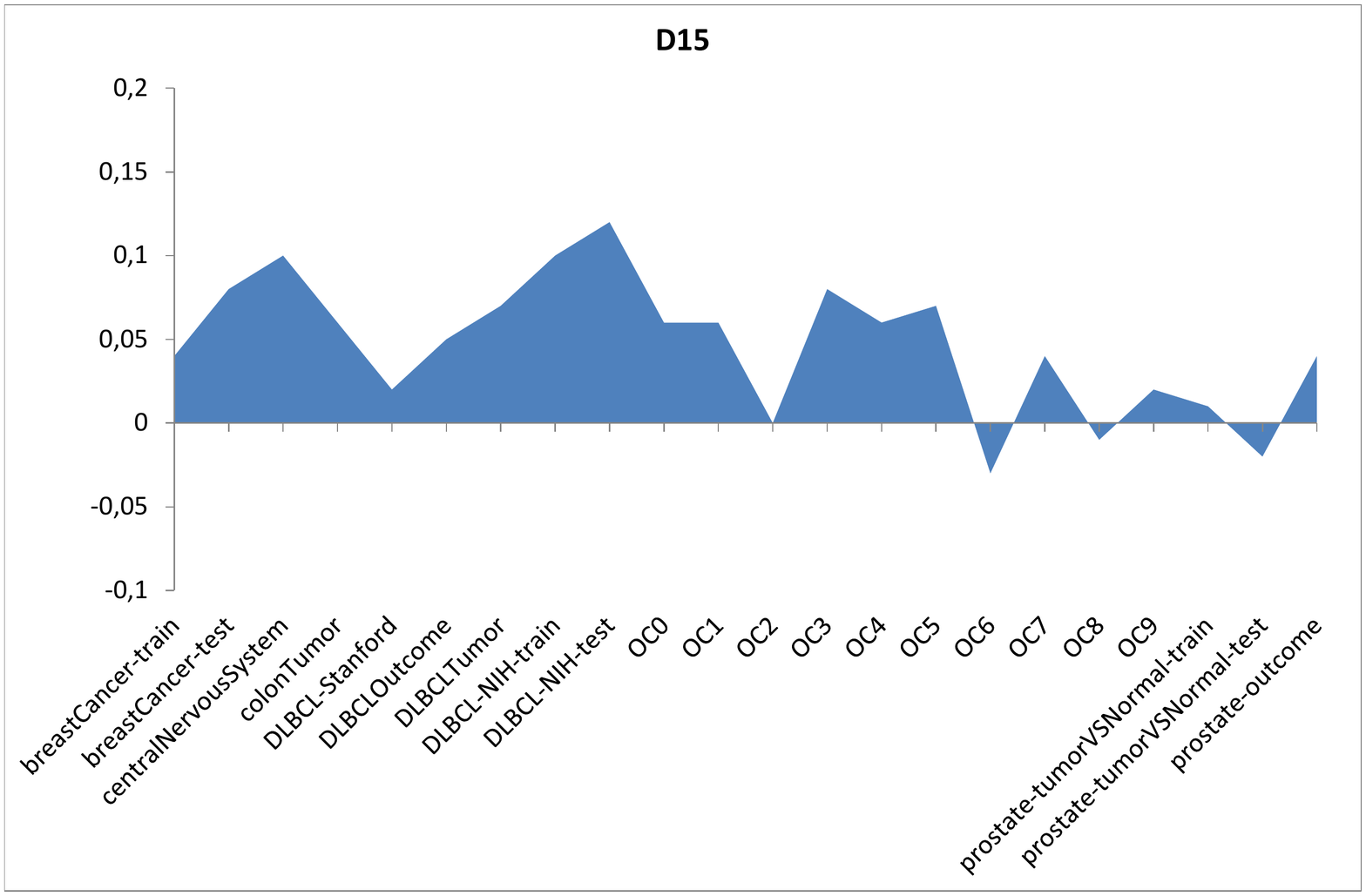}
	\caption{Comparison of the D15 ensemble and the best performing classifiers from Naive Bayes, Adaboost, Bagging and Random Forest.}	
	\label{fig:d15}
\end{figure*}

The mean accuracy and their standard deviations on the datasets for the ensembles can be found in table \ref{tab:res}. Each column contains the classification accuracy of the respective ensembles $D_{i}, i \in \{3, 5, 7, 9, 11, 13, 15\}$. The results for the Naive Bayes, Adaboost, Bagging and Random Forest classifiers can be found in table \ref{tab:res_ensemble}. The values in bold for each dataset contain the best performing method. As it can be seen, for each dataset, the proposed approach provides the best values. However, the number of ensembles members varies among the best results. To have a deeper insight on the choice of optimal ensemble size, each investigated ensemble is compared to the best performing among the Naive Bayes, Adaboost, Bagging and Random Forest classifiers. Figures \ref{fig:d3}-\ref{fig:d15} show the difference between the respective ensemble and the best performing other method, where each positive value means that the respective ensemble performed better than the best among the other classifiers, while a negative value shows otherwise. As it can be seen, only the ensembles with 5 and 9 members remain above the other methods all the time. From table \ref{tab:comparison} we can see that the sum of the all differences are the highest for the $D_{5}$ ensemble. That is, based on these experiments, a false labelling ensemble with 5 member can be recommended to generate.  

\begin{figure*}
	\centering
	\includegraphics[width=\linewidth]{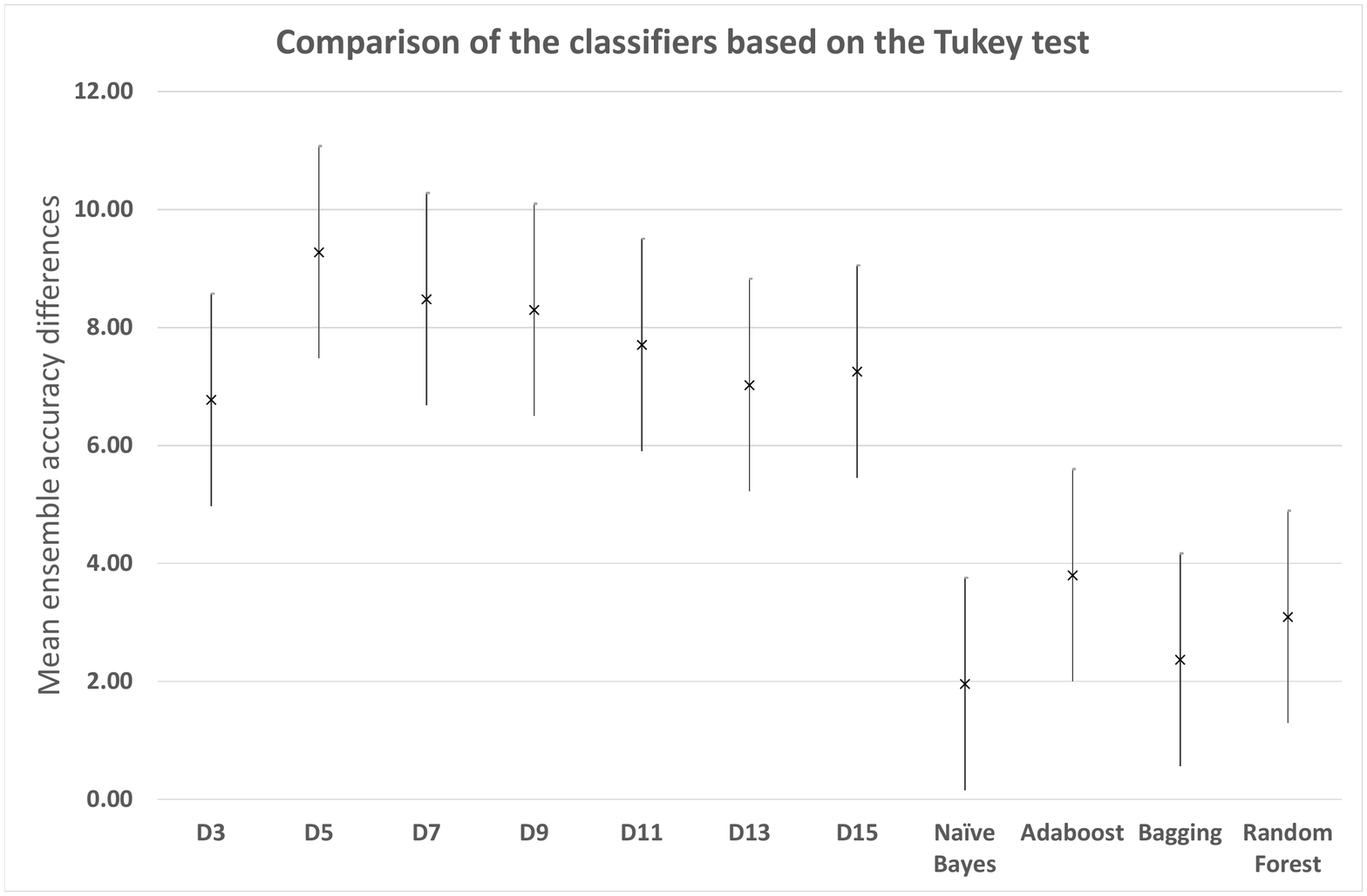}
	\caption{Multiple comparison test}
	\label{fig:tukey}		
\end{figure*}

Statistical analysis of the classifiers is also performed. First, Friedman-test \cite{Friedman} was performed to check whether the results of the proposed ensemble based classifiers, the Naive Bayes, Adaboost, Bagging and Random Forest are from the same distribution. This hypothesis was rejected with p =  3.8499e-026. Then, we applied post-hoc analysis to reveal the differences among the investigated classifiers. To recognize these differences, Tukey's multiple comparison test \cite{Tukey} is also performed. The test revealed that the proposed ensembles consisting of 5-15 member ($D_{5}, \dots, D_{15}$) were all significantly different from the four classifiers they were compared to, while $D_{3}$ were significantly different from all but Adaboost. The Friedman ranking also revealed $D_{5}$ to be the best performing classifier among the investigated ones.  For a visual representation of the Tukey test, see Figure \ref{fig:tukey}, where a confidence interval for the sample mean differences are shown.

\begin{sidewaystable}
\renewcommand{\arraystretch}{1.3}\footnotesize
\caption{Mean and standard deviation of the accuracies of the ensembles on the respective datasets.}
\label{tab:res}
\centering
\begin{tabular}{l c c c c c c c c c}
\hline
Dataset & $D_{3}$  & $D_{5}$  & $D_{7}$  & $D_{9}$  & $D_{11}$  & $D_{13}$  & $D_{15}$\\
\hline

breastCancer-train & 0.88 $\pm$ 0.20 & \textbf{0.98 $\pm$ 0.03} & 0.95 $\pm$ 0.10 & 0.89 $\pm$ 0.14 & 0.88 $\pm$ 0.14 & 0.95 $\pm$ 0.08 & 0.91 $\pm$ 0.16  \\ 

breastCancer-test & 0.98 $\pm$ 0.07 & \textbf{1.00 $\pm$ 0.04} & 0.98 $\pm$ 0.12 & 0.91 $\pm$ 0.25 & 0.92 $\pm$ 0.24 & 0.91 $\pm$ 0.26 & 0.92 $\pm$ 0.24  \\

centralNervousSystem & \textbf{0.98 $\pm$ 0.06} & 0.98 $\pm$ 0.15 & 0.98 $\pm$ 0.11 & 0.96 $\pm$ 0.14 & 0.93 $\pm$ 0.23 & 0.90 $\pm$ 0.23 & 0.94 $\pm$ 0.20  \\

colonTumor & 0.96 $\pm$ 0.09 & \textbf{1.00 $\pm$ 0.04} & 0.99 $\pm$ 0.06 & 0.94 $\pm$ 0.18 & 0.95 $\pm$ 0.16 & 0.95 $\pm$ 0.16 & 0.94 $\pm$ 0.17  \\ 

DLBCL-Stanford & 0.97 $\pm$ 0.08 & \textbf{0.99 $\pm$ 0.10} & 0.98 $\pm$ 0.14 & 0.97 $\pm$ 0.17 & 0.96 $\pm$ 0.20 & 0.94 $\pm$ 0.24 & 0.94 $\pm$ 0.24  \\
DLBCLOutcome & \textbf{0.99 $\pm$ 0.04} & 0.98 $\pm$ 0.14 & 0.99 $\pm$ 0.10 & 0.98 $\pm$ 0.14 & 0.95 $\pm$ 0.22 & 0.92 $\pm$ 0.27 & 0.93 $\pm$ 0.26  \\

DLBCLTumor & 0.98 $\pm$ 0.04 & 0.99 $\pm$ 0.10 & \textbf{1.00 $\pm$ 0.00} & 0.98 $\pm$ 0.14 & 0.98 $\pm$ 0.14 & 0.95 $\pm$ 0.22 & 1.00 $\pm$ 0.00  \\ 
DLBCL-NIH-train & 0.84 $\pm$ 0.05 & 0.87 $\pm$ 0.08 & 0.92 $\pm$ 0.07 & \textbf{0.98 $\pm$ 0.03} & 0.98 $\pm$ 0.07 & 0.92 $\pm$ 0.13 & 0.94 $\pm$ 0.11  \\ 

DLBCL-NIH-test & 0.96 $\pm$ 0.07 & \textbf{1.00 $\pm$ 0.00} & 0.99 $\pm$ 0.10 & 0.97 $\pm$ 0.17 & 0.96 $\pm$ 0.20 & 0.95 $\pm$ 0.22 & 0.94 $\pm$ 0.24  \\

OC0 & 0.91 $\pm$ 0.08 & 0.97 $\pm$ 0.05 & 0.96 $\pm$ 0.05 & \textbf{0.98 $\pm$ 0.03} & 0.97 $\pm$ 0.07 & 0.96 $\pm$ 0.12 & 0.98 $\pm$ 0.03  \\

OC1 & 0.88 $\pm$ 0.06 & 0.93 $\pm$ 0.05 & 0.95 $\pm$ 0.04 & 0.94 $\pm$ 0.07 & 0.91 $\pm$ 0.12 & \textbf{0.98 $\pm$ 0.02} & 0.95 $\pm$ 0.12  \\

OC2 & 0.88 $\pm$ 0.05 & \textbf{0.95 $\pm$ 0.05} & 0.94 $\pm$ 0.06 & 0.92 $\pm$ 0.11 & 0.95 $\pm$ 0.05 & 0.92 $\pm$ 0.11 & 0.87 $\pm$ 0.13  \\

OC3 & 0.90 $\pm$ 0.05 & 0.92 $\pm$ 0.06 & 0.88 $\pm$ 0.11 & 0.93 $\pm$ 0.11 & 0.93 $\pm$ 0.09 & 0.93 $\pm$ 0.10 & 0.96 $\pm$ 0.04  \\

OC4 & 0.92 $\pm$ 0.07 & 0.95 $\pm$ 0.08 & 0.91 $\pm$ 0.09 & 0.93 $\pm$ 0.10 & 0.95 $\pm$ 0.08 & 0.92 $\pm$ 0.10 & \textbf{0.98 $\pm$ 0.04}  \\

OC5 & 0.95 $\pm$ 0.05 & 0.96 $\pm$ 0.06 & 0.95 $\pm$ 0.08 & \textbf{0.98 $\pm$ 0.02} & 0.96 $\pm$ 0.09 & 0.94 $\pm$ 0.09 & 0.94 $\pm$ 0.08  \\

OC6 & 0.92 $\pm$ 0.06 &\textbf{0.95 $\pm$ 0.03} & 0.95 $\pm$ 0.06 & 0.92 $\pm$ 0.08 & 0.92 $\pm$ 0.10 & 0.92 $\pm$ 0.08 & 0.89 $\pm$ 0.15  \\

OC7 & 0.93 $\pm$ 0.05 & 0.95 $\pm$ 0.07 & 0.98 $\pm$ 0.04 & 0.97 $\pm$ 0.03 & 0.96 $\pm$ 0.07 & \textbf{0.99 $\pm$ 0.03} & 0.98 $\pm$ 0.04  \\

OC8 & 0.86 $\pm$ 0.09 & 0.93 $\pm$ 0.03 & 0.91 $\pm$ 0.10 & 0.96 $\pm$ 0.05 & 0.93 $\pm$ 0.07 & \textbf{0.97 $\pm$ 0.07} & 0.92 $\pm$ 0.09  \\

OC9 & 0.89 $\pm$ 0.05 & 0.93 $\pm$ 0.05 & 0.92 $\pm$ 0.05 & \textbf{0.94 $\pm$ 0.07} & 0.93 $\pm$ 0.10 & 0.91 $\pm$ 0.08 & 0.90 $\pm$ 0.06  \\ 

prostate-tumorVSNormal-train &  0.91 $\pm$ 0.14 & \textbf{1.00 $\pm$ 0.00} & \textbf{1.00 $\pm$ 0.00} & \textbf{1.00 $\pm$ 0.00} & 0.95 $\pm$ 0.20 & 0.99 $\pm$ 0.07 & 0.95 $\pm$ 0.18  \\ 

prostate-tumorVSNormal-test &  \textbf{1.00 $\pm$ 0.02} & 0.98 $\pm$ 0.09 & 0.98 $\pm$ 0.10 & 0.96 $\pm$ 0.16 & 0.93 $\pm$ 0.17 & 0.90 $\pm$ 0.21 & 0.92 $\pm$ 0.17\\

prostate-outcome & \textbf{1.00 $\pm$ 0.05} & 0.98 $\pm$ 0.09 & 0.91 $\pm$ 0.27 & 0.90 $\pm$ 0.26 & 0.92 $\pm$ 0.22 & 0.91 $\pm$ 0.24 & 0.93 $\pm$ 0.22  \\

\hline
\end{tabular}   
\end{sidewaystable}

\begin{sidewaystable}\footnotesize
\renewcommand{\arraystretch}{1.3}
\caption{Mean and standard deviation of the accuracies of other state-of-the-art classifiers on the respective datasets.}
\label{tab:res_ensemble}
\centering
\begin{tabular}{l c c c c}
\hline
Dataset & Naive Bayes & Adaboost & Bagging & Random Forest \\
\hline
breastCancer-train & 0.83 $\pm$ 0.10 & 0.87 $\pm$ 0.10 & 0.74 $\pm$ 0.17 & 0.87 $\pm$ 0.08\\
breastCancer-test & 0.84 $\pm$ 0.15 & 0.78 $\pm$ 0.08 & 0.74 $\pm$ 0.15 & 0.80 $\pm$ 0.12\\
centralNervousSystem & 0.83 $\pm$ 0.07 & 0.84 $\pm$ 0.10 & 0.78 $\pm$ 0.16 & 0.81 $\pm$ 0.11\\
colonTumor & 0.84 $\pm$ 0.09 & 0.86 $\pm$ 0.09 & 0.88 $\pm$ 0.14 & 0.86 $\pm$ 0.09 \\
DLBCL-Stanford & 0.92 $\pm$ 0.07 & 0.82 $\pm$ 0.09 & 0.78 $\pm$ 0.12 & 0.88 $\pm$ 0.08\\
DLBCLOutcome & 0.77 $\pm$ 0.19 & 0.86 $\pm$ 0.10 & 0.85 $\pm$ 0.17 & 0.88 $\pm$ 0.13\\
DLBCLTumor & 0.92 $\pm$ 0.08 & 0.93 $\pm$ 0.05 & 0.89 $\pm$ 0.13 & 0.93 $\pm$ 0.08\\
DLBCL-NIH-train & 0.79 $\pm$ 0.08 & 0.83 $\pm$ 0.10 & 0.78 $\pm$ 0.11 & 0.84 $\pm$ 0.10\\
DLBCL-NIH-test & 0.82 $\pm$ 0.13 & 0.77 $\pm$ 0.10 & 0.76 $\pm$ 0.10 & 0.80 $\pm$ 0.17\\
OC0 & 0.82 $\pm$ 0.08 & 0.92 $\pm$ 0.07 & 0.91 $\pm$ 0.04 & 0.89 $\pm$ 0.01 \\
OC1 & 0.81 $\pm$ 0.09 & 0.87 $\pm$ 0.05 & 0.89 $\pm$ 0.04 & 0.86 $\pm$ 0.08 \\
OC2 & 0.81 $\pm$ 0.02 & 0.86 $\pm$ 0.08 & 0.87 $\pm$ 0.05 & 0.84 $\pm$ 0.12 \\
OC3 & 0.85 $\pm$ 0.08 & 0.88 $\pm$ 0.09 & 0.84 $\pm$ 0.07 & 0.85 $\pm$ 0.09 \\
OC4 & 0.83 $\pm$ 0.07 & 0.92 $\pm$ 0.06 & 0.91 $\pm$ 0.05 & 0.91 $\pm$ 0.08 \\
OC5 & 0.85 $\pm$ 0.06 & 0.86 $\pm$ 0.07 & 0.86 $\pm$ 0.07 & 0.87 $\pm$ 0.06 \\
OC6 & 0.88 $\pm$ 0.03 & 0.92 $\pm$ 0.03 & 0.90 $\pm$ 0.02 & 0.86 $\pm$ 0.07 \\
OC7 & 0.81 $\pm$ 0.08 & 0.94 $\pm$ 0.06 & 0.92 $\pm$ 0.04 & 0.92 $\pm$ 0.05 \\
OC8 & 0.85 $\pm$ 0.06 & 0.89 $\pm$ 0.08 & 0.90 $\pm$ 0.06 & 0.93 $\pm$ 0.05 \\
OC9 & 0.84 $\pm$ 0.03 & 0.88 $\pm$ 0.06 & 0.88 $\pm$ 0.08 & 0.87 $\pm$ 0.06 \\
prostate-tumorVSNormal-train & 0.62 $\pm$ 0.05 & 0.94 $\pm$ 0.03 & 0.88 $\pm$ 0.05 & 0.87 $\pm$ 0.03 \\
prostate-tumorVSNormal-test & 0.92 $\pm$ 0.07 & 0.94 $\pm$ 0.06 & 0.87 $\pm$ 0.11 & 0.94 $\pm$ 0.04 \\
prostate-outcome & 0.87 $\pm$ 0.10 & 0.89 $\pm$ 0.12 & 0.78 $\pm$ 0.23 & 0.78 $\pm$ 0.10 \\
\hline

\end{tabular}   
\end{sidewaystable}

\begin{table*}\footnotesize
\renewcommand{\arraystretch}{1.3}
\caption{Difference of the respective ensembles and the best performing methods from table \ref{tab:res_ensemble}.}
\label{tab:comparison}
\centering
\begin{tabular}{l c c c c c c c}
\hline
Dataset &D3&D5&D7&D9&D11&D13&D15\\
\hline
breastCancer-train&0.01&0.11&0.08&0.02&0.01&0.08&0.04\\
breastCancer-test&0.14&0.16&0.14&0.07&0.08&0.07&0.08\\
centralNervousSystem&0.14&0.14&0.14&0.12&0.09&0.06&0.1\\
colonTumor&0.08&0.12&0.11&0.06&0.07&0.07&0.06\\
DLBCL-Stanford&0.05&0.07&0.06&0.05&0.04&0.02&0.02\\
DLBCLOutcome&0.11&0.1&0.11&0.1&0.07&0.04&0.05\\
DLBCLTumor&0.05&0.06&0.07&0.05&0.05&0.02&0.07\\
DLBCL-NIH-train&0&0.03&0.08&0.14&0.14&0.08&0.1\\
DLBCL-NIH-test&0.14&0.18&0.17&0.15&0.14&0.13&0.12\\
OC0&-0.01&0.05&0.04&0.06&0.05&0.04&0.06\\
OC1&-0.01&0.04&0.06&0.05&0.02&0.09&0.06\\
OC2&0.01&0.08&0.07&0.05&0.08&0.05&0\\
OC3&0.02&0.04&0&0.05&0.05&0.05&0.08\\
OC4&0&0.03&-0.01&0.01&0.03&0&0.06\\
OC5&0.08&0.09&0.08&0.11&0.09&0.07&0.07\\
OC6&0&0.03&0.03&0&0&0&-0.03\\
OC7&-0.01&0.01&0.04&0.03&0.02&0.05&0.04\\
OC8&-0.07&0&-0.02&0.03&0&0.04&-0.01\\
OC9&0.01&0.05&0.04&0.06&0.05&0.03&0.02\\
prostate-tumorVSNormal-train&-0.03&0.06&0.06&0.06&0.01&0.05&0.01\\
prostate-tumorVSNormal-test&0.06&0.04&0.04&0.02&-0.01&-0.04&-0.02\\
prostate-outcome&0.11&0.09&0.02&0.01&0.03&0.02&0.04\\
\hline
sum&0.88&\textbf{1.58}&1.41&1.3&1.11&1.02&1.02\\
\hline
\end{tabular}   
\end{table*}

\section{Conclusion}
\label{sec:conclusion}

In this paper, a novel classifier ensemble creation approach is presented. The presented approach automatically creates an2 optimal labelling for a number of classifiers based on the output of a classifier, which are then assigned to the original data instances and fed to classifiers. The approach has been evaluated on high-dimensional biomedical datasets and compared to state-of-the-art classifiers. The results shown improvement in classification accuracy. The possible ensemble size is also investigated, with having 5 ensemble members as an accurate choice. 
The presented approach is the first ensemble creation algorithm which creates diversity among classifiers using an artificially created labelling, a technique which can hopefully reused to create more robust algorithms in problems where individual classifier accuracy can be very varying. 
In the future, the ensemble creation method could be extended to handle unbalanced or multiclass classification problems efficiently. 

\section*{Acknowledgments}
The publication was supported by the T\'AMOP-4.2.2.C-11/1/KONV-2012-0001 project. The project has been supported by the European Union, co-financed by the European Social Fund.

\bibliographystyle{elsarticle-num} 
\bibliography{refs} 


\balance
%
%
\end{document}